\documentclass[times,onecolumn,final,authoryear]{elsarticle}

\usepackage{amssymb}
\usepackage{latexsym}

\usepackage{url}
\usepackage{xcolor}
\usepackage{subfigure}
\usepackage{amsmath}
\usepackage{color}
\usepackage{pifont}
\usepackage[colorlinks,linkcolor=magenta]{hyperref}

\usepackage{geometry}
\definecolor{newcolor}{rgb}{.8,.349,.1}

\begin{document}

\begin{frontmatter}

\title{YOWO-Plus: An Incremental Improvement}

\author[1]{Jianhua Yang}

\ead{19B908049@stu.hit.edu.cn}

\address[1]{State Key Laboratory of Robotics and System, Harbin Institute of Technology, Harbin
150001, China}

\begin{abstract}
  In this technical report, we would like to introduce our updates to YOWO, a real-time method for spatio-temporal
  action detection. We make a bunch of little design changes to make it better. For network structure, we use
  the same ones of official implemented YOWO, including 3D-ResNext-101 and YOLOv2, but we use a better pretrained
  weight of our reimplemented YOLOv2, which is better than the official YOLOv2. We also optimize the label assignment
  used in YOWO. To accurately detection action instances, we deploy GIoU loss for box regression. After our incremental
  improvement, YOWO achieves 84.9\% frame mAP and 50.5\% video mAP on the UCF101-24, significantly higher than the official
  YOWO. On the AVA, our optimized YOWO achieves 20.6\% frame mAP with 16 frames, also exceeding the official YOWO. With
  32 frames, our YOWO achieves 21.6 frame mAP with 25 FPS on an RTX 3090 GPU. We name the optimized YOWO as YOWO-Plus.
  Moreover, we replace the 3D-ResNext-101 with the efficient 3D-ShuffleNet-v2 to design a lightweight action detector,
  YOWO-Nano. YOWO-Nano achieves 81.0 \% frame mAP and 49.7\% video frame mAP with over 90 FPS on the UCF101-24. It also
  achieves 18.4 \% frame mAP with about 90 FPS on the AVA. As far as we know, YOWO-Nano is the fastest state-of-the-art
  action detector. Our code is available on \url{https://github.com/yjh0410/PyTorch_YOWO}.
\end{abstract}

\begin{keyword}
  Spatio-temporal action detection \sep You Only Watch Once \sep real-time action detector

\end{keyword}

\end{frontmatter}

%\linenumbers

%% main text
\section{Introduction}
\label{introduction}
Spatio-temporal action detection (STAD) is a fundamental and important challenge in video understanding. It aims to 
detect action instances in the current input frame. It has been widely applied, such as video surveillance\citep{clapes2018action}
and somatosensory game\citep{yan2019stat}.

Recently, I have taken an interest in STAD, so I'm going to reimplement a popular model to get familiar with this task. Since
YOWO\citep{kopuklu2019you} is simple and efficient, I decided to reproduce it. To my surprise, I found that my reimplemented
YOWO performs better. So, after reimplementing, I released my code, and I'm writing this technical report to tell you what
improvements we make to YOWO, how we do it, and what performance we achieve.

\section{Our improvement}
\label{approach}
\begin{figure}[]
  \centering
  \includegraphics[width=1.0\linewidth]{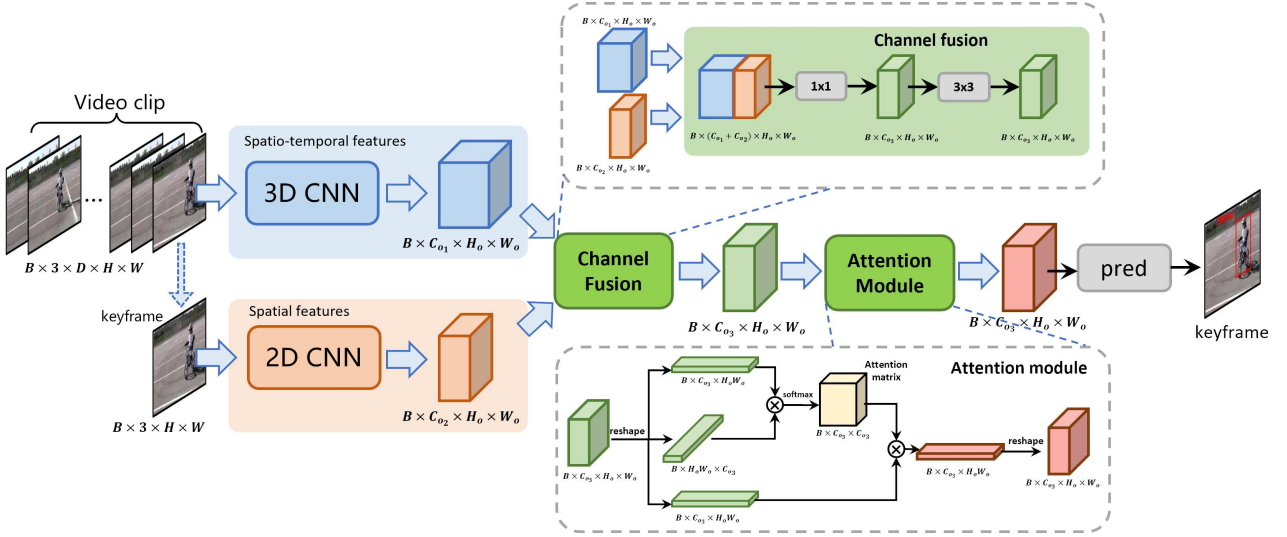}
  \caption{Overview of YOWO.}
  \label{fig:yowo}
\end{figure}

In this technical report, we are dealing with a real-time action detector, YOWO\citep{kopuklu2019you}. It consists of a 3D
backbone and a 2D backbone. YOWO designs a CFAM (channel fusion \& attention module) to fuse 2D spatio-temporal features and
3D spatio-temporal features. After that, it uses a convolutional layer to make the prediction. The whole pipeline of YOWO is shown
in Fig.\ref{fig:yowo}. YOWO claims that it is the fastest state-of-the-art method in spatio-temporal action detection. However,
we think there is still large room for improvement. We optimize YOWO from three aspects: backbone, label assignment, and loss function.

\subsection{Better backbone}
YOWO deploys two backbones to process the input video clip. The 2D backbone only processes the current frame a.k.a keyframe to extract
spatial features, while the 3D backbone processes the whole video clip to extract spatio-temporal features. For the 3D backbone,
YOWO uses the 3D-ResNext-101\citep{kopuklu2019resource} which is pretrained on the Kinetics dataset\citep{carreira2017quo}. Since
it is time-consuming to train a 3D CNN on the Kinetics, we keep the pretrained weight of 3D-ResNext-101. For the 2D backbone,
we use a better COCO pretrained weight of our reimplemented YOLOv2\footnote[1]{\url{https://github.com/yjh0410/PyTorch_YOLOv2}}, 
which achieves 27\% mAP with input $416\times 416$ on the COCO, significantly better than official YOLOv2\citep{redmon2017yolo9000}.
Other structure keeps the same as YOWO.

\subsection{Better label assignment}
YOWO takes advantage of YOLOv2 to process the keyframe and uses the same label assignment as YOLOv2. Given a groundtruth, YOWO
first calculates the grid cell coordinate $\left(grid_x, grid_y\right)$ as shown in Eq.(\ref{eq:grid_cell}),
\begin{equation}
    grid_x = \lfloor \frac{c_x}{s}\rfloor, \ \  grid_y = \lfloor \frac{c_y}{s}\rfloor
  \label{eq:grid_cell}
\end{equation}
where $\left(c_x, c_y\right)$ is the center point of the groundtruth, and $s$ is the output stride of the YOLOv2. Then YOWO
calculates the IoU of the 5 predicted bounding boxes here. Only the predicted box with the highest IoU is used to calculate 
the confidence loss, classification loss, and box regression loss.

Different from the YOWO, we calculate the IoU of the 5 anchor boxes, not the predicted boxes, and then we assign the anchor boxes
with IoU higher than 0.5 to the groundtruth. Therefore, a groundtruth might be assigned with multiple positive samples.

\subsection{Better loss function}
We define loss function as follows:
\begin{equation}
  \begin{aligned}
      L = \ & \lambda_{act}\sum_{i=0}^{S^2}\sum_{j=0}^{B}I_{ij}^{act}\left(C_i - \hat{C}_i\right)^2  + \lambda_{noact}\sum_{i=0}^{S^2}\sum_{j=0}^{B}I_{ij}^{noact}\left(C_i - \hat{C}_i\right)^2 \\
            & + \lambda_{cls}\sum_{i=0}^{S^2}\sum_{j=0}^{B}I_{ij}^{act}FocalLoss \left(p_i, \hat{p}_i\right) \\
            & + \lambda_{coord}\sum_{i=0}^{S^2}\sum_{j=0}^{B}I_{ij}^{act}GIoU \left(b_i, \hat{b}_i\right) \\
    \end{aligned}\label{eq:loss}
\end{equation}
where $\lambda_{act}=5.0$, $\lambda_{noact}=1.0$, $\lambda_{cls}=1.0$ and $\lambda_{coord}$. $I_{ij}^{act}$ and $I_{ij}^{noact}$
are the indicator functions. We use the same confidence loss and classification loss as YOWO does except for the box regression loss.
For box regression, we use the GIoU loss\citep{rezatofighi2019generalized} rather than the smooth L1 loss\citep{ren2015faster}
used in YOWO. During the training phase, the loss is normalized by the batch size.

\subsection{YOWO-Nano}
To design a lightweight action detector, we replace the 3D-ResNext-101 with 3D-ShuffleNet-v2\citep{kopuklu2019resource}.
Other configurations are the same as those of YOWO-Plus. We name this very efficient detector as \textbf{YOWO-Nano}. 

\section{Experiments}\label{experiments}
\subsection{Datasets}
\textbf{UCF101-24}\citep{soomro2012ucf101}. UCF101-24 contains 3,207 untrimmed videos for 24 sports classes
and provides corresponding spatio-temporal annotations. There may be multiple action instances per frame. Following
YOWO\citep{kopuklu2019you}, we train and evaluate YOWO-Plus on the first split.

\textbf{AVA}\cite{gu2018ava}. AVA is a large-scale benchmark for spatial-temporal action detection. It contains
430 15-minute video clips with 80 atomic visual actions (AVA). It provides annotations at 1 Hz in space
and time, and precise spatio-temporal annotations with possibly multiple annotations for each person. Therefore,
this benchmark is very challenging. Following YOWO, we train YOWO-Plus on the train split and evaluate YOWO-Plus on the
most-frequent 60 action classes of the AVA dataset. We report evaluation results on the AVA v2.2.

\subsection{Implementation details}\label{subsection:train_cfg}
For training, we use AdamW optimizer with an initial learning rate 0.0001 and weight decay 0.0005. On the UCF101-24, we
train YOWO-Plus for 5 epochs and decay the learning rate by a factor of 10 at 1, 2, 3, and 4 epoch, respectively. On the AVA,
we train YOWO-Plus for 10 epochs and decay the learning rate by a factor of 10 at 3, 4, 5, and 6 epoch, respectively.
Unless otherwise specified, the size of the input frame is reshaped to $224 \times 224$.

\begin{table}[]
  \caption{Comparison with YOWO on the UCF101-24. FPS si measured on a GPU RTX 3090. K is the length of the video clip.}
  \centering
  \resizebox{0.8\linewidth}{!}{
  \begin{tabular}{c|c|c|c|c|c|c}
  \hline
  Method      & K  & FPS   & F-mAP (\%)  & V-mAP (\%)  & FLOPs       & Params      \\ \hline\hline
  YOWO        & 16 & 34    & 80.4        & 48.8        & 43.7 B       & 121.4 M     \\
  YOWO+LFB    & -  & -     & 87.3        & 53.1        & -            & -           \\ \hline
  YOWO-Plus   & 16 & 35    & 84.9        & 50.5        & 43.7 B       & 121.4 M     \\
  YOWO-Nano   & 16 & 91    & 81.0        & 49.7        & 6.0  B       & 72.7 M      \\ \hline
  \end{tabular}}
  \label{comparison_yowo}
  \end{table}

\begin{table}[]
  \caption{Comparison with YOWO on the AVA. FPS si measured on a GPU RTX 3090.}
  \centering
  \resizebox{0.35\linewidth}{!}{
  \begin{tabular}{c|c|c|c}
  \hline
  Method      & K  & FPS   & mAP    \\ \hline\hline
  YOWO        & 16 & 31    & 17.9   \\
  YOWO        & 32 & 23    & 19.1   \\
  YOWO+LFB    & -  & -     & 20.2   \\ \hline
  YOWO-Plus   & 16 & 33    & 20.6   \\
  YOWO-Plus   & 32 & 25    & 21.6   \\
  YOWO-Nano   & 16 & 91    & 18.4  \\
  YOWO-Nano   & 16 & 90    & 19.5  \\ \hline
  \end{tabular}}
  \label{comparison_yowo_ava}
  \end{table}

\subsection{Comparison with YOWO}\label{subsection:comparison_with_yowo}
\textbf{UCF101-24}. Table.\ref{comparison_yowo} summarizes the results on the UCF101-24. From the table, YOWO-Plus achieves
better performance than YOWO (84.9\% frame mAP v.s. 80.4 \% frame mAP). Only equipped with long-term feature bank (LFB)\citep{wu2019long},
YOWO can surpass YOWO-Plus. However, YOWO equipped with the LFB is no longer a causal model because it uses the future
information for inference, so it can only run offline. Moreover, equipped with 3D-ShuffleNet-v2, YOWO-Nano achieves better
a trade-off between performance and detection speed, exceeding YOWO with the higher mAP and much fewer FLOPs.
YOWO-Nano can run at 91 FPS, satisfying the real-time detection requirements. 

\textbf{AVA}. Table.\ref{comparison_yowo_ava} summarizes the results on the AVA. Both YOWO-Plus and YOWO-Nano surpass YOWO
with stronger performance and faster detection speed, demonstrating the effectiveness of our improvements.

Although plenty of ablation experiments should be made to prove the effectiveness of our improvements, it is too much
time-consuming. I could only finish the design of YOWO-Plus and YOWO-Nano in my spare time. I just want to honestly share
some of the improvements I've made to YOWO, not to publish a paper. So, I think that this technical report can be ended here.
If you are interested in YOWO-Plus and YOWO-Nano, please try our code in GitHub and star it. I hope you will enjoy it.
If you have any questions, feel free to email me or leave an issue on github.

\section{Acknowledgment}
Big thanks to YOWO\footnote[2]{https://github.com/wei-tim/YOWO} for designing such an excellent action detector and releasing
their code.

\bibliographystyle{model2-names}
\bibliography{refs}

\end{document}